\documentclass[utf8]{frontiersSCNS} 

\usepackage{hyperref}
\usepackage{lineno}
\usepackage{subcaption}
\usepackage[onehalfspacing]{setspace}

\usepackage{amssymb}
\usepackage{color}
\usepackage{lineno}

\usepackage[margin=1.1in,a4paper]{geometry}
\usepackage{booktabs}
\usepackage{caption}
\usepackage{float}
\usepackage{titlesec}
\usepackage{capt-of}
\usepackage{multirow}
\usepackage{tabularx}
\usepackage{array}
\usepackage{arydshln}
\usepackage{subcaption}
\usepackage{caption}
\usepackage{graphicx}
\usepackage{subcaption}
\usepackage{float}
\usepackage{booktabs}
\usepackage{todonotes}

\def\keyFont{\fontsize{8}{11}\helveticabold }
\def\firstAuthorLast{Zhou Yang {et~al.}} 
\def\Authors{Zhou Yang\,$^{1}$, Edward Dougherty\,$^{2}$, Chen Zhang\,$^{3}$, Zhenhe Pan\,$^{3}$, and Fang Jin\,$^{1*}$}
\def\Address{$^{1}$Department of Statistics, George Washington University, Washington D.C., USA\\$^{2}$Department of Mathematics, Roger Williams University, Rhode Island, USA\\
$^{3}$Department of Computer Science, Texas Tech University, Texas, USA }
\def\corrAuthor{Fang Jin}

\def\corrEmail{fangjin@gwu.edu}

\address{United States}

\begin{document}
\onecolumn \firstpage{1}

\title[Discovering Transmission Dynamics of COVID-19 in China]{Discovering
  Transmission Dynamics of COVID-19 in China}

\author[\firstAuthorLast ]{\Authors} 
\address{} 
\correspondance{} 

\extraAuth{}

\maketitle


\begin{abstract}

  A comprehensive retrospective analysis of public health interventions
  including large-scale testing, quarantining and in-depth contract tracing,
  that have been implemented nationwide to ease and contain the transmission of
  the novel SARS-CoV-2 virus, would help identify and comprehend those
  mechanisms most successful in the mitigation of the COVID-19 pandemic. In this
  work, we comprehensively investigate China-based transmission patterns of the SARS-CoV-2
  virus, such as infection type and potential transmission source, through
  publicly release tracking data using an assemblage of data processing and
  quantitative assessment techniques. To begin, we exhaustively mine these data,
  that include reported infection cases, from multiple sources including local
  health commissions, the Chinese Center for Disease Control and Prevention, and
  official social media outlets of local governments. The collected data are
  then processed using Natural Language Processing methods and collated by our
  research team to formulate a transmission/tracking chain. In addition, a
  robust statistical analysis of the tracking data in conjunction with
  population mobility data from the city of Wuhan is implemented to help
  quantify and visualize both temporal and spatial dynamics of the COVID-19
  disease spread. Results show that transmission differs greatly among different
  geographic regions, with larger cities having more infections due to
  traditional social activities. In addition, the vast majority $(79\%)$ of
  individuals that were symptomatic were admitted to the hospital within just 5
  days of symptom presentation, while those individuals that had contact to
  confirmed cases self-admitted to a hospital in less than 5 days. Further,
  results demonstrate that the type of infection source evolved over time; at
  the onset of outbreak, the majority of COVID-19 cases were directly associated
  with travel to or a contact link to travel to the Hubei Province, while
  transmission became more linked to social activities as the pandemic
  progressed.

  \tiny \keyFont{ \section{Keywords:} COVID-19, Transmission Dynamics, Case
    Study Statistics, Epidemiological
    Progession} 
\end{abstract}

\section{Introduction}
\label{S:1}
As of July 1, 2020, the outbreak of Coronavirus Disease 2019 (COVID-19) has been
contained in the mainland of China. This conclusion is supported by the fact
that the number of new daily reported cases is less than 50
~\footnote{http://www.gov.cn/xinwen/2020-05/02/content\_5508222.htm}. Since late
January 2020, the recognized onset of COVID-19, a series of public health
interventions such as hierarchical lock-downs~\citep{mei2020policy,yang2020covid19},
large-scale population testing ~\citep{xin2020negative}, quarantining and case
by case contact and infection tracking ~\citep{day2020covid} have been
implemented across China to help control and contain the spread of this
epidemic.

As with any infectious disease epidemic, understanding the infection dynamics of
COVID-19 is a crucial step for accurately determining appropriate mitigation
strategies and for planning medical and treatment resources as the disease
circulation accelerates. In addition, the acquired knowledge facilitates
preparation and planning for combating future epidemic and pandemic challenges.
In this light, the wide-spread use of the aforementioned Chinese interventions
provides a rich data source for developing a greater understanding of COVID-19
transmission tendencies and dynamics. In fact, with an appropriate collection of
analytical approaches can be used to provide a comprehensive retrospective
analysis that enables an assessment of COVID-19 infection dissemination as well
as intervention efficacy.

The analysis techniques presented in this work embody a multi-step, hierarchical
approach, leveraging methods from data mining and processing as well as
statistical models~\cite{nguyen2019forecasting}. Confirmed COVID-19 cases from 33 Provinces and regions of
China are collected, from numerous public sources including news outlets,
governmental agencies, social media posts, and local health commissions~\cite{yang2017harvey, du2019twitter}. Then,
demographics, travel, lifestyle, and social interaction information of these
confirmed cases are extracted, organized, and stored. Natural Language
Processing methods in conjunction manually probing formulated an annotation of
the positive cases, which extends the data to also include information including
travel specific to Wuhan, Hubei, disease incubation period, subsequent
infections rendered, time period between infection and presentation of symptoms,
etc. Then, as a means for beginning to understand infection sources and disease
dynamics, infection types are classified and categorize. Finally, relationships
among demographics and transmission aspects are established to quantify
correlations between these variables. It is at this point that trends and higher
level characteristics of infection sources, transmission types, and mitigation
strategies for COVID-19 are assessed, providing unique and practical information
on the dynamics of this disease.

To our knowledge, the approaches presented in this paper provide the most
in-depth quantitative assessment of confirmed case demographics, lifestyle, and
population contact from publicly available data for the COVID-19 pandemic in
mainland China; while non-trivial, they provide greater utility for drawing
conclusions about disease dynamics that encompass both individual behaviors and
higher-level population trends. Compare this outcome to, for example, the common
practice of making a static case by case disease transmission or a simplistic
case transmission ratio, crudely obtained by just dividing the number of cases
in each transmission type can be misleading. Case by case analyses are
well-known to lack the ability to take into consideration, or even provide
information for, overall disease dissemination trends; their disadvantages
extend to an inability to unveil greater incidence characters throughout the
greater population as they operate at just the individual level. Furthermore,
transmission ratios completely miss fine-grained patterns as they do not possess
the fidelity to incorporate important individual-level data. The methods of this
paper overcome these limitations by providing the ability to synthesize the
case-level analysis with higher-level transmissions characteristics.
Specifically, results will expose symptom associations, age distribution,
preparedness of medical facilities, population traveling pathways, infection
geographic distributions and spatio-temporal dynamics of disease transmission.

\section{Methods}
\label{S:2}

\subsection{Data}
Data of confirmed COVID-19 cases was collected from 33 provinces and regions in
China, excluding the Hubei province, for a total of $4,899$ records. Sources of
this information included newspapers, local news channels, national news
organizations (CCTV), China's CDC website, official social media accounts, and
local health commissions. These data were generated as reports, wither when an
individual has a positive COVID-19 test result or when they were admitted to a
hospital. These reports are presented in a retrospective narrative format, and
possess information including travel history, individuals of which they had
close social contacted, recent lifestyle activities, infectious status of family
members, data of symptom onset, existing chronic conditions, etc.

Our team developed a web crawler to comprehensively collect the full report of
each of these confirmed cases ~\citep{leung2020first, yang2020coordinating}, which were then
transferred into JSON format and archived in our custom repository, which we
refer to as a Trajectory Database. Though the reports do not conform to a
standardized template, they do possess a significant amount of shared patient
information including travel history, gender, age, summarized daily mobility,
symptoms, and chronic disease. In addition, the following information is also
retained in the Trajectory Database when available: dates of COVID-19 symptom
onset, travel specific to Wuhan, Hubei or other cities, travel details,
epidemiological links to confirmed COVID-19 cases with details including
relationships, contact times, and events, residential district, predicted date
of exposure, date of symptom onset when applicable, date of hospital admission,
and the time between the date of symptom onset and date of positive COVID-19
result or date of admission to hospital. These collated reports are the
foundation of the Trajectory Database.

\subsection{Data Processing and Annotation}
The $4,899$ were manually mined and annotated to extract and organize key
patient information features including infection type (i.e. source), infection
media, incubation period, number of people they transmitted the COVID-19 virus
to, and common transmission locations. Specifically, for each confirmed case,
data annotations include: (i) demographics including age and gender, (ii)
residence location, (iii) travel history, including travel to Wuhan, Hubei or
neither, (iv) infection type which could be Hubei related travel, social
transmission, relatives, or public transit, (v) social transmission locations
including hospital, supermarket, or restaurant, (vi) calculated incubation
period, (vii) number of transmissions they initiated, (viii) relationship to
initiated transmissions, (ix) specific individuals whom they had contact with
and whether these individuals are a confirmed case, (x) presentation of a
chronic disease(s), (xi) confirmed date of infection, and (xii) date of severe
COVID-19 symptom presentation if applicable. To achieve accurate annotations, we
manually annotated all collected reports; a blinded cross-evaluation validation
process was implemented, where the same reports was assigned to more than one
team member. Annotations were stored in the Trajectory Database only when a
blinded unanimous consensus was achieved, or when an additional team member
confirmed annotation results. The team included one professor and seven graduate
studetnts, and by assessing the cross-evaluation validation approach, annotation
accuracy exceeded 96\%.

\subsection{Infection Types}
Infection source, herein referred to as type, is classified into four main
categories, namely travel to Hubei, public transit, social activities, and
relative. It is well-known that the COVID-19 outbreak commenced in Wuhan, Hubei;
thus ``infected by traveling to Hubei" serves as a main source of infection, and
dissemination to other provinces is naturally followed. Public transit including
bus, train, and metro system, incorporates the majority of the population's
typical daily transportation activities. Public transit in particular provides
wide-spread mobility of, and interaction by, individuals confirmed COVID-19
positive, thereby providing a vehicle ideal for rapid disease spread. The
Chinese populous is rooted in close ties of relatives, making infection via
relative transmission a substantial source. Finally, social transmission is a
well-recognized community-based transmission type, which we identify with common
locations including shopping malls, restaurants, hospitals, residences,
supermarkets, hotels, and nursing home facilities. A compelling advantage of
categorizing cases in this manner is that a high-level summary of overall
transmission characteristics can be acquired, if a researcher desires, without
involving specific case-by-case details.

We employed the pre-trained Bidirectional Encoder Representations from
Transformers (BERT) ~\citep{devlin2018bert,chang2019x} model to classify the
infection type of each report, based upon the report details. We refer to
~\citep{devlin2018bert,chang2019x} for a detailed exposition of the BERT model
and provide an overview here as it relates to our work. Given a training data
set $\mathcal{D}=\left\{\left(\mathbf{x}_{i}, \mathbf{y}_{i}\right) \mid
  \mathbf{x}_{i} \in \mathcal{X}, \mathbf{y}_{i} \in\right.$ $\left.\{0,1\}^{L},
  i=1, \ldots, N\right\}$, where $ \mathbf{x}_{i} \in \mathcal{X}$ is the input,
$\mathbf{y}_{i}$ is the label, $N$ is the number of the input, and $L$ denotes
the number of classes in a label $\mathbf{y}_{i}$, a multi-class text
classification aims to learn a scoring function $f$ that maps an input
$\mathbf{x}_{i}$ to a score $f(\mathbf{x}_{i}) \in \mathbb{R}$, such that the
the total misclassification loss $L(\mathbf{y}, f(\mathbf{x}))$ is minimized. A
simplified representation of the problem is
\begin{align}
  \mathrm{f}^{*} =\underset{\mathrm{f}}{\arg \min } L(\mathbf{y}, \mathrm{f}(\mathbf{x})),
  \label{equation:optimal_f}
\end{align}



where $f^*$ is the desired learned scoring function. With the BERT model, we
first classify the four main transmission types, namely ``infected by social
activities'', ``infected by public transit'', ``infected by relative'', and
``infected by travelling to Hubei''. Then details of each transmission type are
examined in further detail by categorizing them into travel-related and
community-related, the former of which is then further refined into public
transportation, relative transportation, and social transportation. For
community-related infection transmission, sub-categories include
relative-related, public transit-related, and social-related, with each group
forming the following additional sub-groups: infection via vehicle, restaurant,
bus, train, supermarket, airport, hotel, shopping mall, and residential. We note
that a classification of ''unknonwn" is listed when a sub-grouping can not be
confidently established.

\subsection{Gradient Boosting Regression}
\label{section:GBR}
The Gradient Boosting Regression (GBR) algorithm was customized to extrapolate
the relationship amoung case report variables. For example, GBR was used for the
relationship between ``reported cases'' and the ``distance from Wuhan'', as well
as the percentage of outflow of population from Wuhan to a particular city and
the corresponding ``reported cases''. This approach is formalized as the
following regression problem:

Given a training set ${D}= \left\{\left(\mathbf{x}_{i},
    y_{i}\right)\right\}_{i=1}^{n}$, where $\mathbf{x}_{i} \in R^{D}$ is the
input and $y_{i} \in R$ is the true value, find a function $\mathrm{f}$ that
maps $\mathbf{x}$ to $\mathbf{y}$ while minimizing the specified loss function
$L(\mathbf{y}, \mathrm{f}(\mathbf{x}))$ given by Equation
[\ref{equation:optimal_f}]. Note in this formalization that
$\mathbf{x}=\left\{\mathbf{x}_{i}\right\}_{i=1}^{n}$ is then a D-dimensional
(input) matrix with dimension $D \times n$, and
$\mathbf{y}=\left\{y_{i}\right\}_{i=1}^{n}$ is a one dimensional (target) vector
of length n.

The GBR approach approximates $\mathrm{f}^{*}$ in an additive expansion form of
$f$. That is, $f$ is approximates by $\mathrm{f}_M(\mathbf{x})=\sum_{m=0}^{M}
\beta_{m} h_{m}(\mathbf{x})$, where $h_{m}(\mathbf{x})$ are the base learners,
and $M$ is the number of iterations. GBR starts with a constant value
$f_{0}(x)=\underset{\gamma}{\arg \min } \sum_{i=1}^{n} L\left(y_{i},
  \gamma\right)$, where $\gamma$ is the parameters to be learned. For each
iteration $m$, the model computes the so-called pseudo-residuals
$y_{i,m}=-\left[\frac{\partial L\left(y_{i},
      f\left(x_{i}\right)\right)}{\partial f\left(x_{i}\right)}\right]$, where
${f(x)=f_{m-1}(x)}$ holds for $i=1, \ldots, n$. Then, a base learner $h_{m}(x)$
is fit to pseudo-residuals by training it via the training set
$\left\{\left(x_{i}, r_{i m}\right)\right\}_{i=1}^{n}$; $f_m$ is updated by
$f_{m}(x)=f_{m-1}(x)+\gamma_{m} h_{m}(x)$, where the multiplier $\gamma_{m}$ is
computed by solving $\gamma_{m}=\underset{\gamma}{\arg \min } \sum_{i=1}^{n}
L\left(y_{i}, F_{m-1}\left(x_{i}\right)+\gamma h_{m}\left(x_{i}\right)\right)$.
The final $f_{M}$ in this process is the desired, i.e. optimal, $f^*$ in
Equation~[\ref{equation:optimal_f}]. Related spatio-temporal learning and scheduling problems have also been studied in other domains, motivating transferable modeling choices for complex dynamic systems~\citep{nguyen2019spatial,nguyen2019forecasting, liang2020data}.

\section{Infection dynamics Analysis}
\begin{figure}[t]
  \centering
  \includegraphics[width=0.58\linewidth]{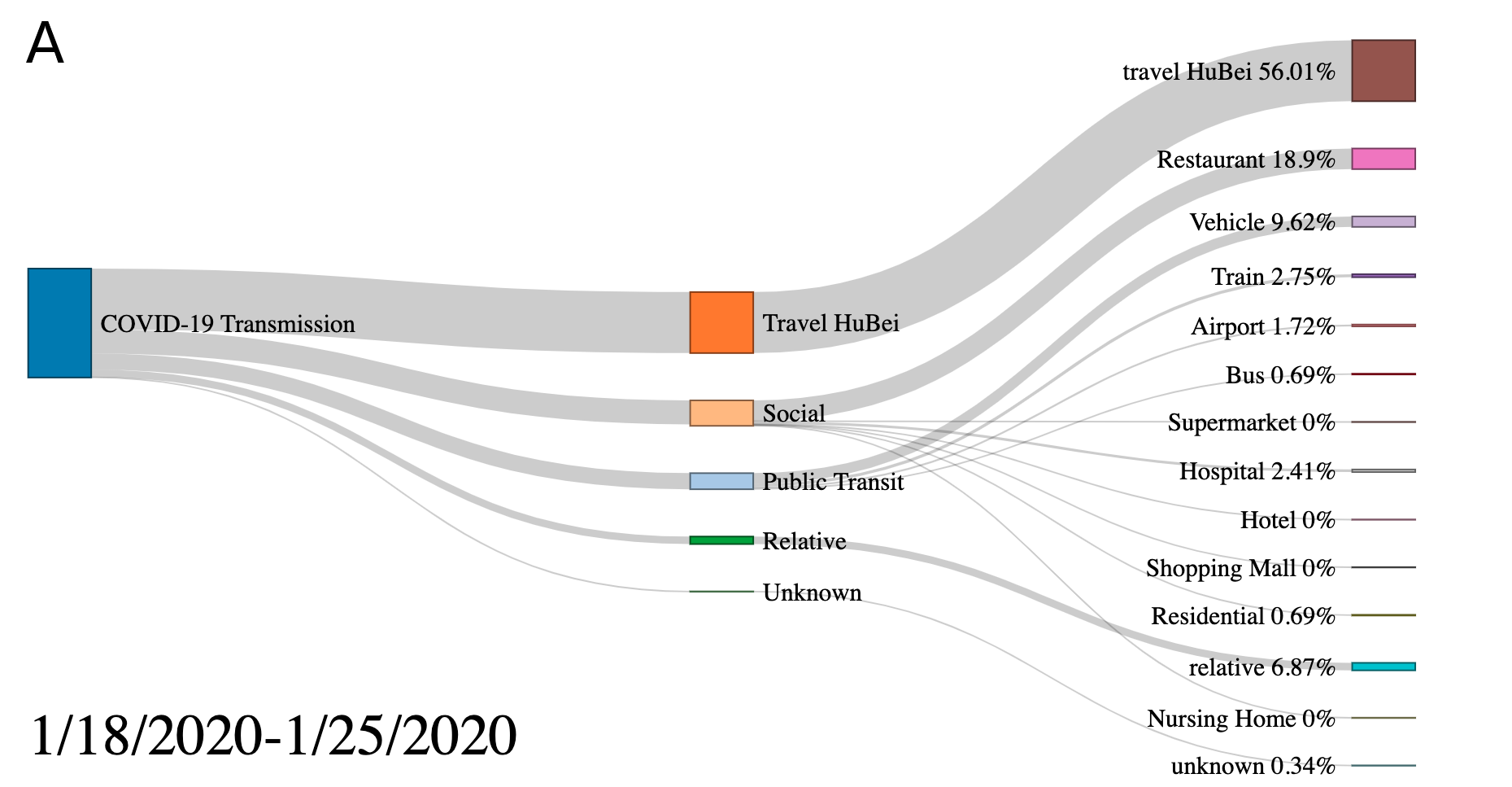}
  \includegraphics[width=0.58\linewidth]{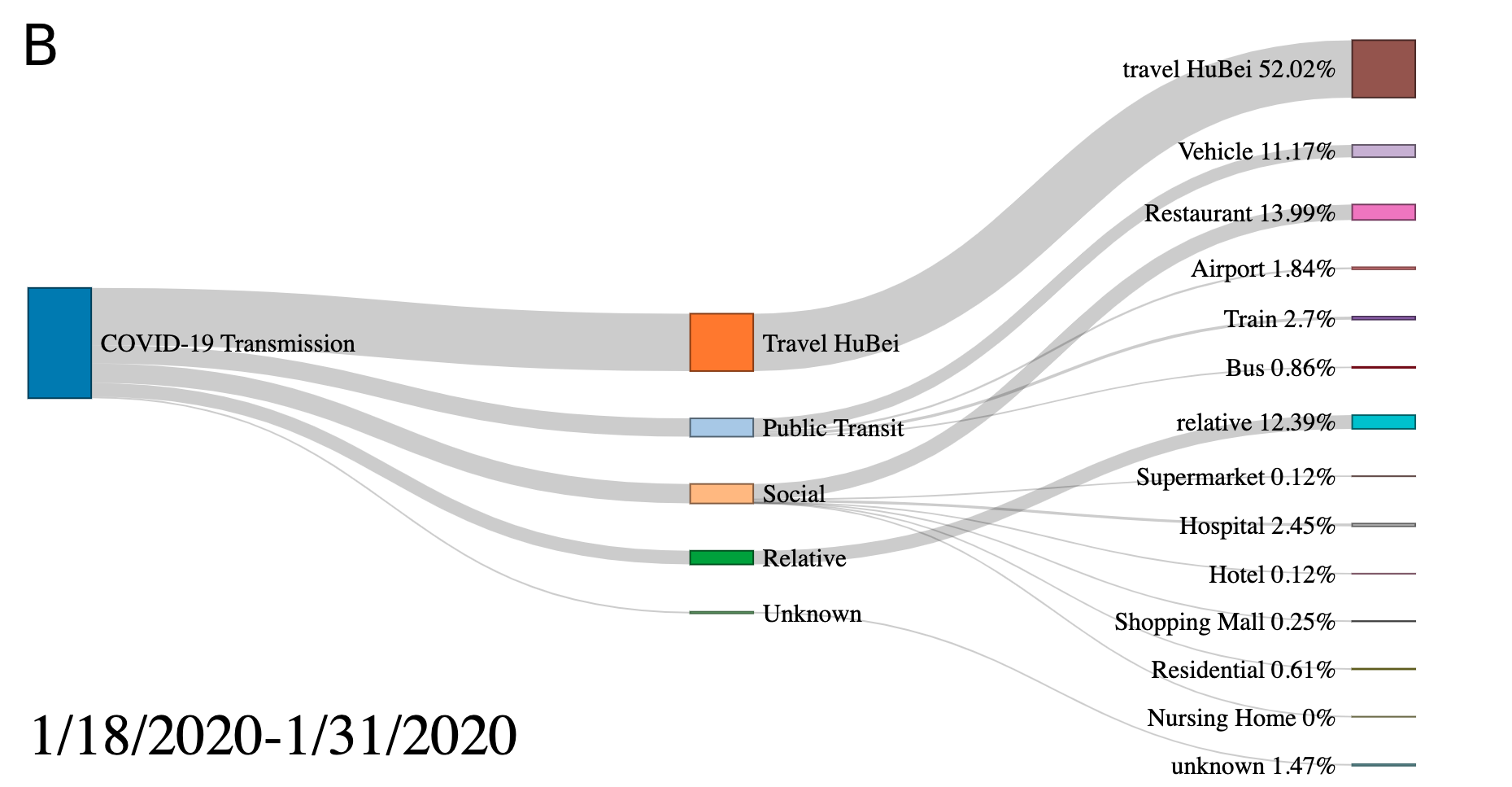}
  \includegraphics[width=0.58\linewidth]{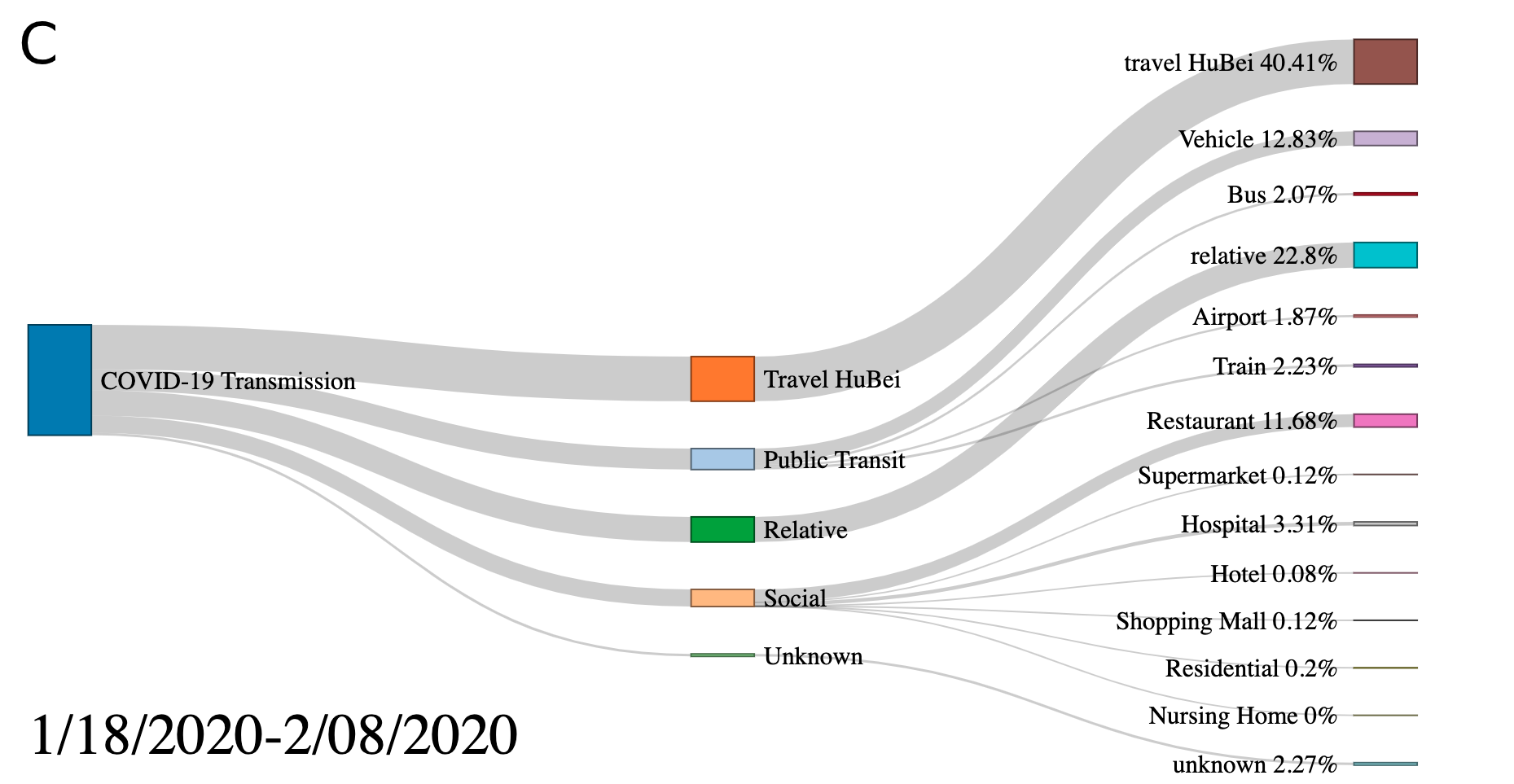}
  \includegraphics[width=0.58\linewidth]{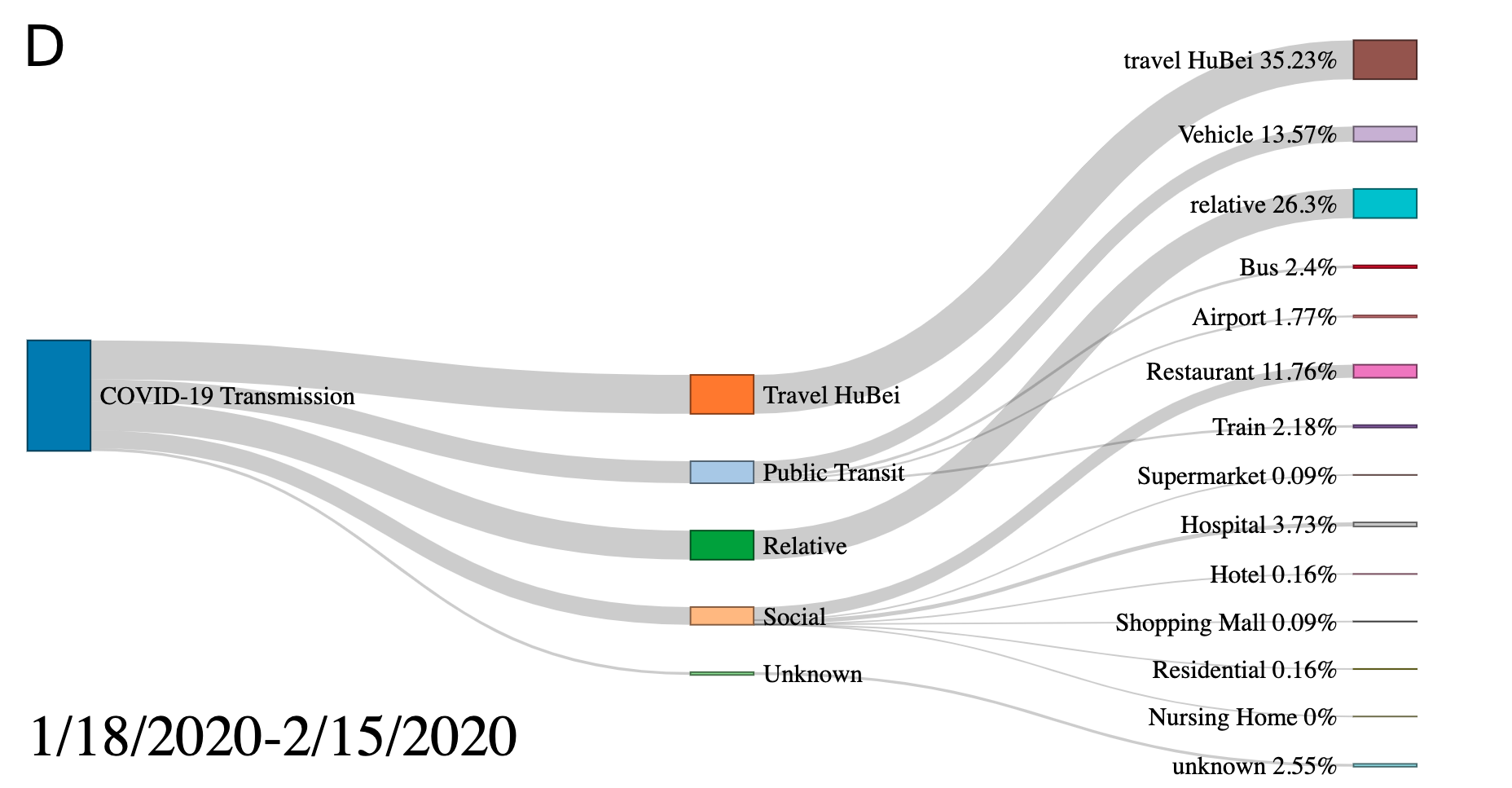}
  \includegraphics[width=0.58\linewidth]{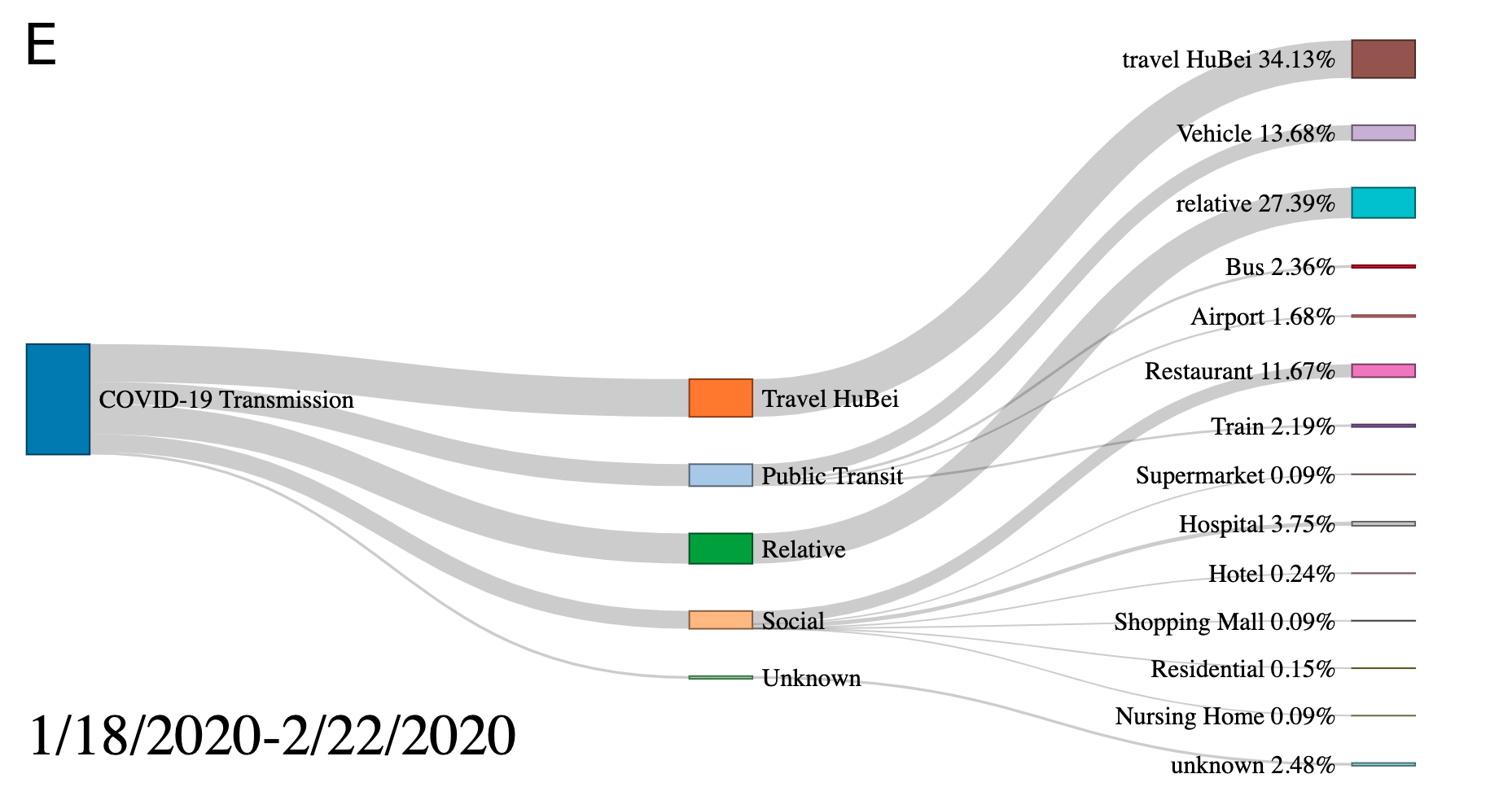}
  \caption{Cumulative reported cases tracing back to infection types in 5 weeks.
  }
  \label{fig:week15}
\end{figure}

A comprehensive assessment of the dynamics of each infection type is presented
in this section. Infection by nature is a dynamic process, and so model results
are analyzed within a weekly time bin result weekly to unveil micro and macro
characteristics of the infection types. The infection types classification is
illustrated in Figure~\ref{fig:week15}, and shows the hierarchical
classification percentage and the temporal difference among different period;
the latter is derived by comparing figures from the corresponding period. In
addition, Table~\ref{tab:types} presents the numerical summary of transmission
dynamics of each infection type sub-category. Here, we compare the dynamics of
infection types extracted from the tracking dataset, and summarize the infection
dynamics of ``traveling to Hubei'', ``public transit'', ``social'', and
``relative'' and the corresponding sub-dynamics into Temporal Transmission
Dynamics and Spatial Transmission Dynamics. In the following subsections, we
present a detailed analysis of the aforementioned data, table and figures.

\begin{table*}[htbp]
  \centering
  \caption{Weekly transmission dynamics and percentages of each subcategory. }
  \begin{small}
    \begin{tabular}{lllrrrrr}
      \toprule
      Categories    &   Subcat.  & Sub-subcat.      & 1 Week & 2 Weeks & 3 Weeks & 4 Weeks & 5 Weeks \\
      \midrule
      \multicolumn{1}{l}{\multirow{12}[6]{*}{\shortstack{Local \\ Trans.}}} & \multirow{7}[2]{*}{Soical} & Restaurant & 18.90\% & 13.99\% & 11.68\% & 11.76\% & 11.67\% \\
                    &       & Supermarket & 0\%   & 0.12\% & 0.12\% & 0.09\% & 0.09\% \\
                    &       & Hospital & 2.41\% & 2.45\% & 3.31\% & 3.73\% & 3.75\% \\
                    &       & Hotel & 0\%   & 0.12\% & 0.08\% & 0.16\% & 0.24\% \\
                    &       & Shopping Mall & 0\%   & 0.25\% & 0.12\% & 0.09\% & 0.09\% \\
                    &       & Residential & 0.69\% & 0.61\% & 0.20\% & 0.16\% & 0.15\% \\
                    &       & Nursing Home & 0\%   & 0\%   & 0\%   & 0\%   & 0.09\% \\
      \cmidrule{2-8}          & \multicolumn{1}{l}{\multirow{4}[2]{*}{\shortstack{Public \\ Transit}}} & Private Vehicle & 9.62\% & 11.17\% & 12.83\% & 13.57\% & 13.68\% \\
                    &       & Train & 2.75\% & 2.70\% & 2.23\% & 2.18\% & 2.19\% \\
                    &       & Airport & 1.72\% & 1.84\% & 1.87\% & 1.70\% & 1.68\% \\
                    &       & Bus   & 0.69\% & 0.86\% & 2.07\% & 2.40\% & 2.36\% \\
      \cmidrule{2-8}          & Relative & relative & 6.78\% & 12.39\% & 22.80\% & 26.30\% & 27.39\% \\
      \midrule
      Hubei & Hubei & Hubei & 56.01\% & 55.02\% & 40.41\% & 35.23\% & 34.13\% \\
      \midrule
      Unknown & - & - & 0.34\% & 1.47\% & 2.27\% & 2.55\% & 2.48\% \\
      \bottomrule
    \end{tabular}%
  \end{small}
  \label{tab:types}%
\end{table*}%

\subsection{Temporal Transmission Dynamics} 
Figure~\ref{fig:week15} shows hierarchical transmission dynamics with ratios of
each transmission type changing over time. We first show the main transmission
types, and then we delve into the details of each transmission type in
subsequent sub-figures. As shown in Figure~\ref{fig:week15} A, having a travel
history to Hubei was the main transmission type from January 18, 2020 to January
25, 2020, which accounts for $56.01\%$ of all reported cases. Moreover, infected
by family members, infected on public transportation, and infected in social
activities possesses a similar percentage. Among all the social activities,
dining in restaurants is the main disease transmission type based on
percentages.

At the early stage of the COVID-19 epidemic in China, namely Jan 18 to Jan 25,
2020, the majority of cases (56.01\%) have a travel history to Hubei. This
result is consistent with the official reports and the widely acknowledged
viewpoint that the epicenter of the COVID-19 outbreak is the Hubei province. At
the same time, 18.9\% of the cases are related to restaurants or dining
together, i.e. social activity, with infected individuals. The reason why a
large percentage of the population was infected when dining out is due to the
popular Chinese customs and tradition of dining together with friends and
relatives. For public transportation-related cases, the majority (9.62\%) of
them are traced back to infected by traveling using a private vehicle. One
likely explanation of this is the well-known fact that people attempt to travel
together by private vehicle to avoid the Spring Festival traffic (CITE), which
inevitably increases the chance of becoming infected if a carpool companion is
infected. In this same time period, 6.87\% of the cases are reported to be
infected by relatives. Overall, the vast majority of cases are traced back to
have a travel history to Hubei, and the second most transmission source is
traveling together by private vehicle.

As time passed, these dynamics changed; Figure~\ref{fig:week15} E shows the
transmission statistics from January 18, 2020, to February 22, 2020. Here, it is
observed that 34.13\% of reported cases were traced back to have a travel
history to Hubei province, 27.39\% of reported cases were infected by relatives,
13.68\% of reported cases were infected when traveling by private vehicle, and
11.67\% of confirmed cases were infected in the restaurant. Further,
Table~\ref{tab:types} shows the weekly percentage within each subcategory; as
can be seen in Table~\ref{tab:types}, the percentage of each infection source
subcategory varies as the epidemic progresses. This same observation can be
obtained by comparing the percentages in each subcategory in
Table~\ref{tab:types}. For example, the dominated infection source is observed to be travel history to Hubei, including Wuhan. Yet, as the Hubei province went into lockdown, Hubei-related cases decreased concurrent with an increase in local transmissions.

When comparing the distribution of five weeks' reported cases in
Figure~\ref{fig:week15} and the percentage of each subcategory, we conclude that
there are in fact several dynamical temporal patterns for the transmission
COVID-19 within China. First, having a travel history to the Hubei province is
the main infection source at the early stage with 56.01\%, but the percentage of
this category decreases by 21.88\% to 34.13\%. The decrease is a result of the
lock-down policy and the fact that no people could travel out of Hubei province
after Jan 22, 2020. Second, the percentage of relative-related cases increases
sharply from 6.87\% to 27.39\%. One viable contributor to this observation is
that family members are most likely to infect each other when a stay-at-home
policy is implemented and the community is in a lockeddown state. In this
scenario, even though the stay-at-home policy blocks the spread of the virus, it
magnifies the probability of being infected by family members. Third, the
percentages of cases that trace back to public transportation remains stable in
the train and airplane subcategories. However, the percentage for cases tracing
back to private vehicles expands to 13.68\%, from 9.62\% at the early stage,
while the percentage for cases related to bus transportation increases by 1.67\%
to 2.36\%. Fourth, an obvious decline of 5.92\% in the percentage of cases
tracing back to social activities is observed from the early stage of 22\% to
the later stage of 16.08\%. This decrement is a consequence of the strictly
implemented stay-at-home policy and city lockdown, and it in turn proves the
efficacy of these policies in containing the spread of the virus. Overall, there
is a sharp decline of 21.88\% in the imported cases that have a connection with
Hubei province, whereas the local transmission-based cases climb steadily from
43.7\% to 63.4\%.

\begin{figure}[hbtp]
  \centering \includegraphics[width=\linewidth]{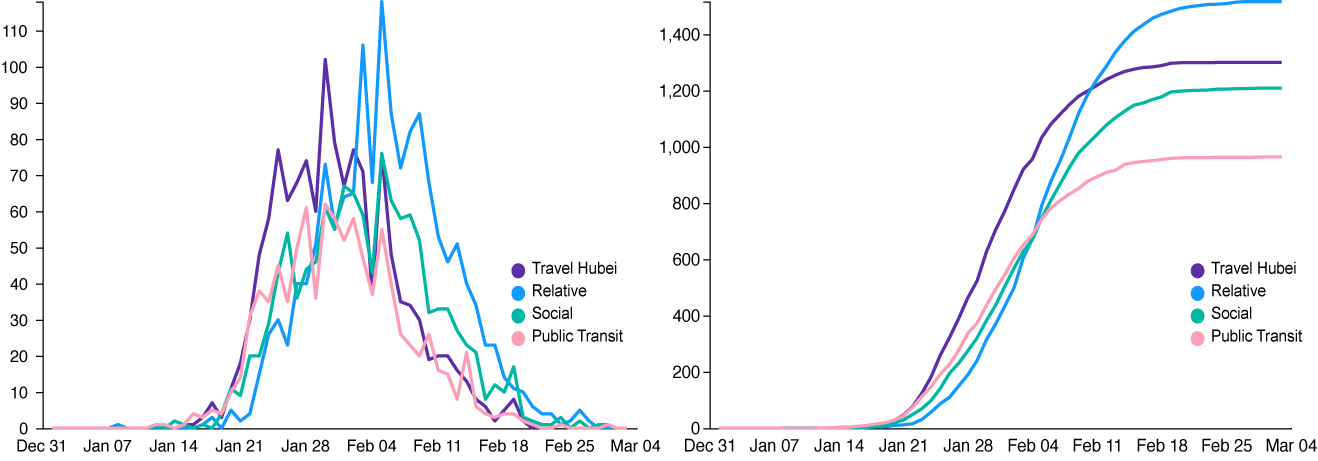}
  \caption{Temporal dynamics of each transmission type. A. Daily reported cases
    for each transmission type. B. Cumulative confirmed cases of each
    transmission type.}
  \label{fig:compare}
\end{figure}

Figure~\ref{fig:compare} illustrates the temporal transmission dynamics of four
categories. Figure~\ref{fig:compare} A shows the daily increment of reported
cases for four subgroups; as we can see, the curves of daily increment related
to public transportation and having a traveling history to Hubei province peak
in February 2020, while the curves of the daily increment that were caused by
close contacts with relatives and social activities both peak on the same day on
February 8, 2020. Also, the date that the curves of daily increment, due to
relative contact and social activities, peak later than the curves of increment
tracing back to a traveling history to Hubei province and public history. This
observation is consistent with the foregoing observations in
Figure~\ref{fig:week15}. Figure~\ref{fig:compare} presents the cumulative cases
of the four groups. Similarly, the increasing trends of cases related to a
traveling history to Hubei and public transportation share the same trend
characteristics, while the increasing tendency of cases related to relative
contacts and social activities are the same. Overall, the key observations in
figure~\ref{fig:compare} give context to and help strengthen the main patterns
and explanations provided by Figure~\ref{fig:week15} and its analys.

\subsection{Spatial Transmission Dynamics}
\begin{figure}[h]
  \centering \includegraphics[width=\linewidth]{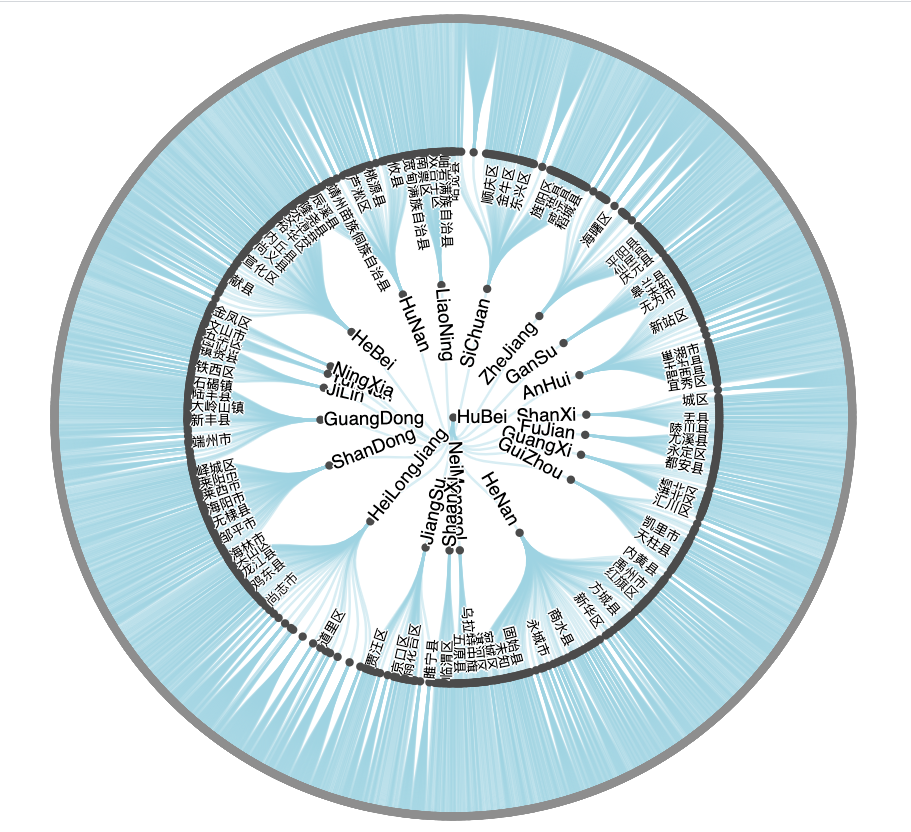}
  \caption{The provincial and municipal distribution of reported cases in the
    tracking dataset. }
  \label{fig:tree}
\end{figure}

The spatial transmission dynamics demonstrate that spatial factors can greatly
influence the distribution of reported cases as well as the geographical
population migration that influences and facilitates the COVID-19 transmission.
For instance, the individuals that were infected but not yet detected did travel
out of Hubei and became the virus spreaders to other provinces and regions.
Thus, an analysis of the population outflow of the Hubei province and the
connection between outflow and reported cases could highlight the patterns that
help explain infection source. The outflow of the reported cases in shown in
Figure~\ref{fig:tree}, and as can be seen, we present two levels of outflow
distribution, the provincial and municipal level.

\begin{figure}[h]
  \centering \includegraphics[width=\linewidth]{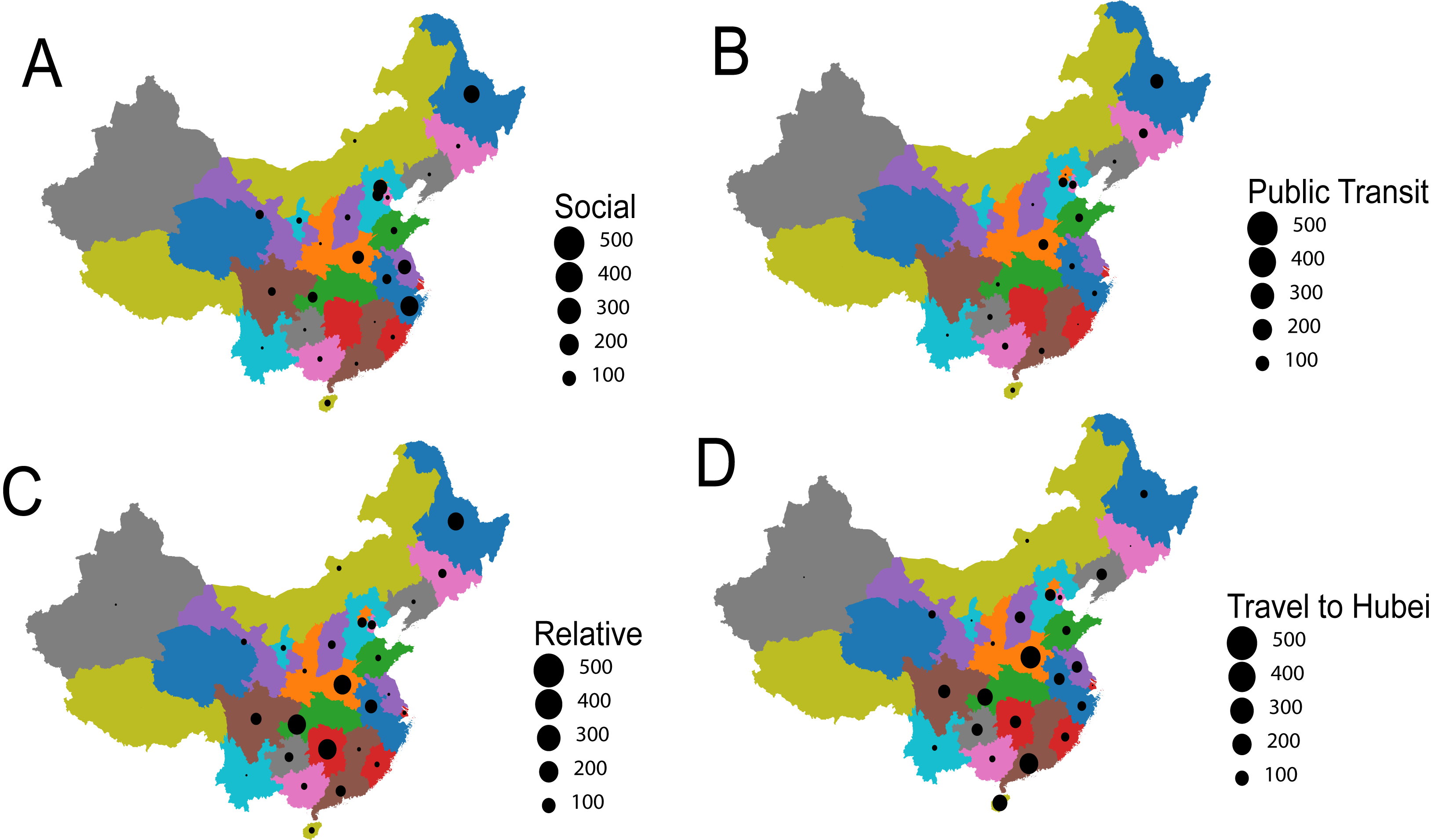}
  \caption{Infection Types Distribution Map in the mainland of China, except
    Hubei province, accumulated case numbers from January 18, 2020, to April 10,
    2020.}
  \label{fig:4map}
\end{figure}

\begin{figure}[h]
  \centering \includegraphics[width=\linewidth]{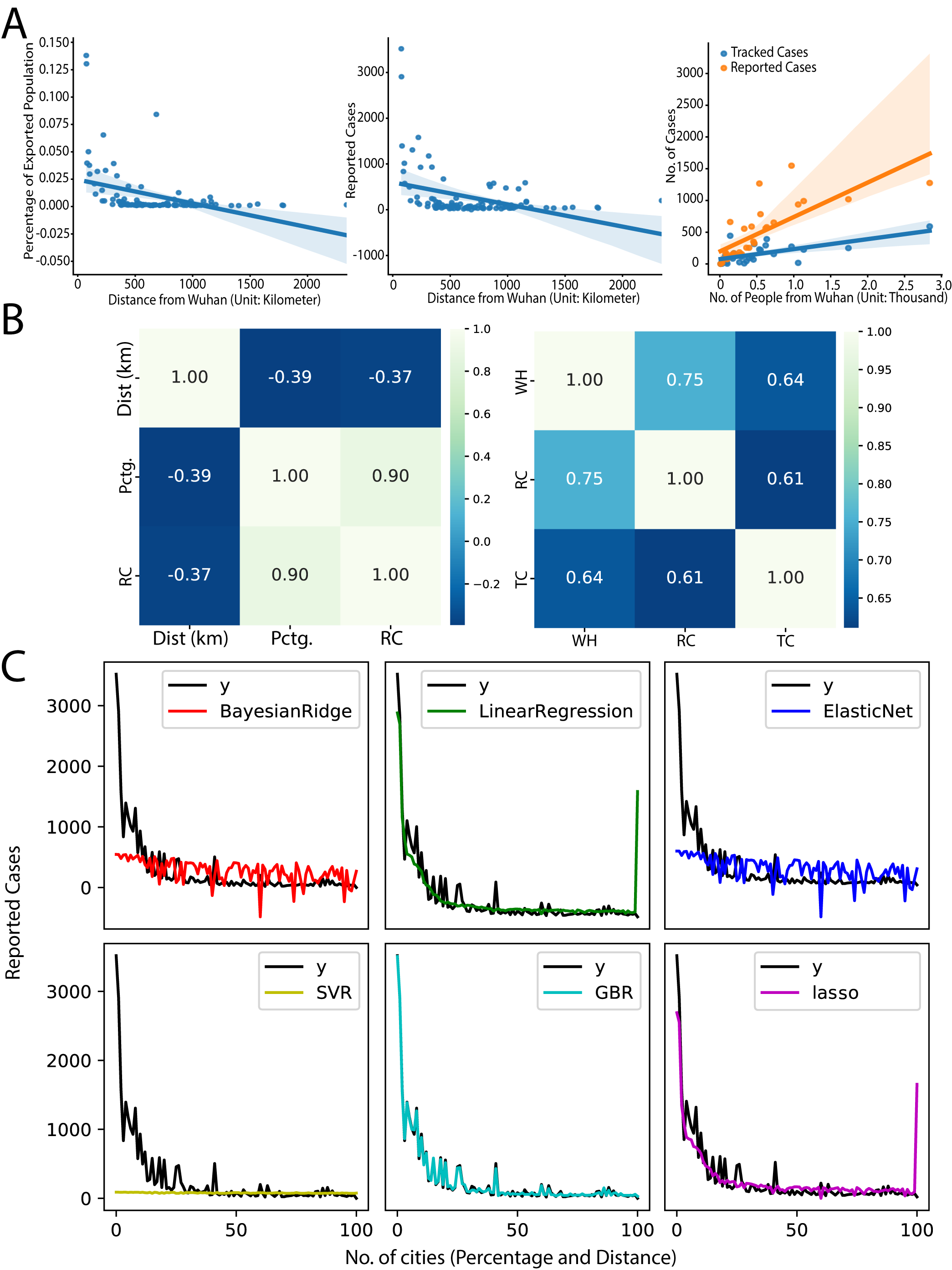}
  \caption{\textbf{A.}The correlation among ``Percentage of Exported Population
    (from Wuhan to top 100 cities)'', ``the Reported Cases'', and the ``Distance
    from Wuhan''. It illustrates that cities that close to Wuhan had more people
    travelling from Wuhan and thus a large number of reported Cases.
    \textbf{B.}The correlation among ``No. of People from Wuahn'', ``Tracked
    Cases'' and ``Reported Cases''. It shows that the more people traveling from
    Wuhan, the more reported cases, and hence the more tracked cases.
    \textbf{C.}Comparisons of regression performance between GBR and baselines.}
  \label{fig:combined_reg}
\end{figure}

Figure~\ref{fig:4map} illustrates the geographic distribution of confirmed cases
that are identified to be infected by social activities, relatives, public
transportation, and a traveling history to the Hubei Province. As shown in
Figure~\ref{fig:4map} A, provinces in the eastern and northeastern regions of
China, including Zhejiang, Jiangsu, Beijing, Henan, and Heilongjiang, have more
confirmed cases that are traced back to infection by social activities. As for
reported cases that connect to public transportation, a province in the coastal
region has more confirmed cases (Figure~\ref{fig:4map} B). Figure~\ref{fig:4map}
C depicts confirmed cases that trace back to be infected by relatives, and as
can be seen, provinces in the central region have more cases when compared to
provinces outside of this region. The distribution pattern of cases that have a
travel history to the Hubei province (Figure~\ref{fig:4map} D) is scattered
around Hubei province, and provinces in the southern region have more confirmed
cases. Also, provinces that are far from the Hubei province have fewer cases
compared to provinces that are geographically closer to Hubei. It is observed
that provinces closest to Hubei, like Henan, have more infections related to
Hubei travel. In general, it is observed in Figure~\ref{fig:4map} that
economically developed provinces have more documented public transit type
infection than less economic developed areas, which can be explained by the
greater presence of public transportation options and infrastructure in these
more affluent provinces. . In addition, relative transmission accounts more in
rural areas than urban cities, highly-likely due to the close familial ties in
these regions. Finally, social transmission appears more uniformly distributed
across provinces, likely explained by the general social actvities and
characteristics of the country as a whole


Besides the differences observed in the spatial/geographic distribution, the
outflow of population from Wuhan and the spatial distance between a city and
Wuhan will also affect the dynamics of the disease transmission. We first
explore the relationship between the number of cases reported and the distance
between Wuhan, and the population out-flowed from Wuhan. This relationship is
visualized in Figure~\ref{fig:combined_reg} A and Figure~\ref{fig:combined_reg}
B. As seen in Figure~\ref{fig:combined_reg} A, the distance from Wuhan and
reported cases is positively correclated and directly proportional. Next, we
explore this relationship with GBR (Section~\ref{section:GBR}), and compare its
performance with popular regression methods including Bayesian Ridge Regression,
Linear Regression, ElasticNet, SVR, and Lasso. These results provide performance
metrics, and are presented in the Table S1 of Supplementary Material. These
results show that GBR consistently outperforms these baselines in terms of Error
Variance, Mean Absolute Error, and Mean Square Error. Performance results are
visualized in Figure~\ref{fig:combined_reg} C, and show further support of the
relationship between the reported cases, and the outflow of population from
Wuhan and the spatial distance between a city and Wuhan.


\section{Discussion}

In this work we implemented a series of statistical analyses using a massive tracking data of
reported cases, and the goal is to understand the transmission dynamics. Such data-driven public-health monitoring can be complemented by personalized, location/time-aware digital interventions~\citep{yang2019addictfree,jayachandra2020besober}. We
summarize several essential observations concerning transmission dynamics from
the perspective age, gender spatial distribution and temporal evolution.

Our results demonstrate that several temporal and spatial patterns of the changing transmission
dynamics of COVID-19 in China. First, having a travel history to Hubei province
is the main infection source at the early stage with 56.01\%, but the percentage
of this category decreases by 21.88\% to 34.13\%. The decrement in percentage is
explainable since Wuhan and even Hubei province was locked down to prevent the
spread of virus. Second, the percentage of relative-related cases increases
sharply from 6.87\% to 27.39\%. One justifiable explanation is that family
members are most likely to infect each other when stay-at-home policy was
implemented and city was locked down. In this scenario, even though stay-at-home
policy blocks the spread of virus, it magnify the probability of being infected
family members. Third, the percentages of cases that trace back to public
transportation keep stable in subcategories such as by train, by airplane. But
the percentage for cases tracing back to private vehicle expands to 13.68\%,
which is 9.62\% at the early stage, while the percentage for cases related to
bus increase by 1.67\% to 2.36\%. Fourth, a obvious decline of 5.92\% in the
percentage of cases tracing back to social activities are observed from the
early stage of 22\% to the later stage of 16.08\%. This decline in statistics is
supported by the strictly implemented stay-at-home policy and city lockdown, and
it in return proves the efficacy of these policy in containing the spread of
virus. Overall, there are a sharp decline of 21.88\% in the imported cases that
have a connection with Hubei province, and the community-based cases climb from
43.7\% to 63.4\%.

\section{Conflict of interest statement}
The authors declare that the research was conducted in the absence of any
commercial or financial relationships that could be construed as a potential
conflict of interest.

\section{Ethics Statement}
This research project was confirmed to be exempt from IRB review and all
personal information is anonymized.

\section{Author contribution statement}
All of the authors Contribute to the paper.

\section{Data Availability Statement}
The datasets GENERATED for this study can be found in the Supplementary
Materials.

\bibliographystyle{frontiersinSCNS_ENG_HUMS} 
\bibliography{reference}

\end{document}